\documentclass[conference]{IEEEtran}
\IEEEoverridecommandlockouts
\usepackage{cite}
\usepackage{amsmath,amssymb,amsfonts}
\usepackage{algorithmic}
\usepackage{graphicx}
\usepackage{textcomp}
\usepackage{xcolor}
\usepackage{latexsym}
\usepackage{subfigure}
\usepackage{times}
\usepackage{soul}
\usepackage{url}
\usepackage[utf8]{inputenc}
\usepackage{caption}

\usepackage{amsthm}
\usepackage[ruled,linesnumbered]{algorithm2e}
\usepackage{amssymb}
\usepackage{multirow}
\usepackage{diagbox}
\usepackage{float}
\usepackage{array}
\usepackage{flushend}
\usepackage{subcaption}
\usepackage{booktabs}
\usepackage{tablefootnote}
\usepackage{wrapfig}
\usepackage{lipsum} % 用于生成示例文本
\usepackage{colortbl}
\usepackage{balance}
\newcolumntype{?}{!{\vrule width 1.2pt}}

\def\BibTeX{{\rm B\kern-.05em{\sc i\kern-.025em b}\kern-.08em
    T\kern-.1667em\lower.7ex\hbox{E}\kern-.125emX}}
\begin{document}

\title{Weighted Graph Structure Learning with Attention Denoising for Node Classification
}

\author{\IEEEauthorblockN{1\textsuperscript{st} Tingting Wang\IEEEauthorrefmark{0\dag}, 2\textsuperscript{th} Jiaxin Su\IEEEauthorrefmark{0\dag}, 3\textsuperscript{nd} Haobing Liu\IEEEauthorrefmark{0*}, 4\textsuperscript{rd} Ruobing Jiang}
\IEEEauthorblockA{\textit{Ocean University of China}\\
Qingdao, China \\
wtt2022@stu.ouc.edu.cn, sujiaxin@stu.ouc.edu.cn, haobingliu@ouc.edu.cn, jrb@ouc.edu.cn}
\thanks{\IEEEauthorrefmark{0\dag} These authors contributed equally to this work and are co-first authors.}
\thanks{\IEEEauthorrefmark{0*} Corresponding author.}
}

\maketitle
\vspace{-0.3cm}
\begin{abstract}
The node classification in graphs aims to predict the categories of unlabeled nodes utilizing a small set of labeled nodes. However, weighted graphs often contain noisy edges and anomalous edge weights, which can distort fine-grained relationships between nodes and hinder accurate classification. We propose the Edge Weight-aware Graph Structure Learning (EWGSL) method, which combines weight learning and graph structure learning to address these issues. EWGSL improves node classification by redefining attention coefficients in graph attention networks to incorporate node features and edge weights. It also applies graph structure learning to sparsify attention coefficients and uses a modified InfoNCE loss function to enhance performance by adapting to denoised graph weights. Extensive experimental results show that EWGSL has an average Micro-F1 improvement of 17.8$\%$ compared to the best baseline.
\end{abstract}

\vspace{-0.2cm}
\section{Introduction}
The primary goal of node classification is to predict the categories of unlabeled nodes using the labels of a small number of known nodes in the graph\cite{refer9}\cite{refer12}. This is a significant problem in graph analysis and finds widespread application in various fields such as social network analysis~\cite{refer25}~\cite{refer26}, and recommendation systems\cite{refer10}. Currently, most approaches employ Graph Neural Networks (GNNs)~\cite{ref3}~\cite{refer28}~\cite{refer7} to leverage graph structure for propagating information and learning node representations for classification\cite{refer13}. Therefore, the quality of the graph structure directly impacts the accuracy of node classification\cite{refer1}\cite{refer3}.

In real-world scenarios, graph structures often contain noisy edges~\cite{ref5}~\cite{ref6}~\cite{refer19} due to uncertainties or erroneous connections between nodes. Noisy edges refer to the edges connecting nodes from different classes, which may lead to errors or misclassifications in node recognition by the model. Common approaches to handling such noise generally fall into two categories~\cite{refer2}. The first category is weight learning methods~\cite{ref14}~\cite{refer7}, which mitigate the impact of noise on the model by adjusting edge weights, and assigning lower weights to noisy edges. The second category is structure learning methods~\cite{ref15}~\cite{refer4}, which aims to reduce noise by optimizing and refining the graph structure to ensure reliable information propagation. However, few studies have integrated both categories simultaneously~\cite{refer6}.

To enhance the ability of the model to handle noisy edges, we combine two approaches to address node classification in weighted graphs, facing two primary challenges. First, edge weights on weighted graphs can describe fine-grained relationships between nodes, with higher weights indicating stronger connections and greater influence on information propagation\cite{refer14}\cite{refer15}. However, existing weight learning methods primarily focus on the structural characteristics of the graph and do not adequately account for edge weights. \textit{How to learn more fine-grained weights between nodes is a challenge.} Second, noisy edges in weighted graphs are characterized by anomalous edge weights. Specific scenarios, such as network failures that affect communication between users for a period of time, can cause edge weights that describe the communication frequency to deviate significantly from normal conditions\cite{refer17}. \textit{How to adaptively correct anomalous edge weights and remove noisy edges is a challenge.}

To address these challenges, we propose an \underline{E}dge \underline{W}eight-aware \underline{G}raph \underline{S}tructure \underline{L}earning (EWGSL) method, a novel model that employs edge weight-aware methods to learn fine-grained weights and employing Graph Structure Learning (GSL)\cite{refer17} to reduce noise interference. Leveraging the excellent scalability of Graph Attention Networks (GATs)\cite{ref4}, we use GAT to learn node representations. We redefine the calculation of attention coefficients in GAT by incorporating both node features and edge weights to capture the relationships between nodes accurately. We apply GSL to sparsify the attention coefficients, setting the attention weights of irrelevant nodes to zero and concentrating them on relevant nodes to effectively identify noisy edges. Additionally, we modify the noise contrastive estimation (InfoNCE) loss function\cite{ref57} to dynamically utilize the weights from the denoised graph structure, thereby enhancing node classification performance. 

\section{Related Work}
\vspace{-0.1cm}
\subsection{Weight Learning Methods}
\vspace{-0.1cm}
Weight learning methods address noisy edges by assigning different weights to the edges, specifically by adjusting the weights of noisy edges to smaller values in order to reduce their impact on node representation updates. In this way, the model enhances the weights of effective edges, ensuring more accurate information propagation and thereby improving the performance of node classification tasks.

WFSM-MaxPWS\cite{refer14} prunes the subgraph expansion process by calculating the ``Maximum Possible Weighted Support'', retaining only subgraphs with higher weighted support. This reduces computational overhead and improves mining efficiency while ensuring the completeness of the results. MetaGC\cite{refer7} adaptively adjusts the weight of each edge in the graph using a meta-model, specifically lowering the weight of noisy edges to mitigate their impact on the clustering results. Additionally, it employs a decomposable clustering loss function and end-to-end training, allowing the method to enhance robustness against noisy edges without relying on external denoising steps.

\vspace{-0.1cm}
\subsection{Structure Learning Methods}
\vspace{-0.1cm}
Structure learning methods handle noisy edges by optimizing the graph structure, primarily through identifying and removing noisy edges, and reconstructing the graph connectivity. This enhances the propagation of information between similar nodes. This approach ensures a more accurate graph structure, improves the quality of node representations, and reduces the interference of noise on model learning.

PTDNet\cite{refer5} employs a parameterized neural network to learn the removal of noisy edges based on the content and structural information of the graph. The network automatically determines which edges are noisy according to the task requirements and enforces sparsity by applying a penalty to the number of edges in the graph. GAM\cite{refer20} utilizes an attention mechanism trained via reinforcement learning to guide the traversal process, focusing only on the task-relevant parts of the graph while ignoring noisy regions, thereby enhancing classification performance and reducing the interference caused by noise.
\vspace{-0.2cm}
\section{The Proposed Model}
\vspace{-0.1cm}
This section begins with definitions relevant to the node classification task, followed by detailed explanations of the two main modules: Edge-aware Weight Learning and Graph Sparsity Structure Learning.

\begin{figure*}[t]
\vspace{-0.38cm}
\centerline{\includegraphics[scale=0.3]{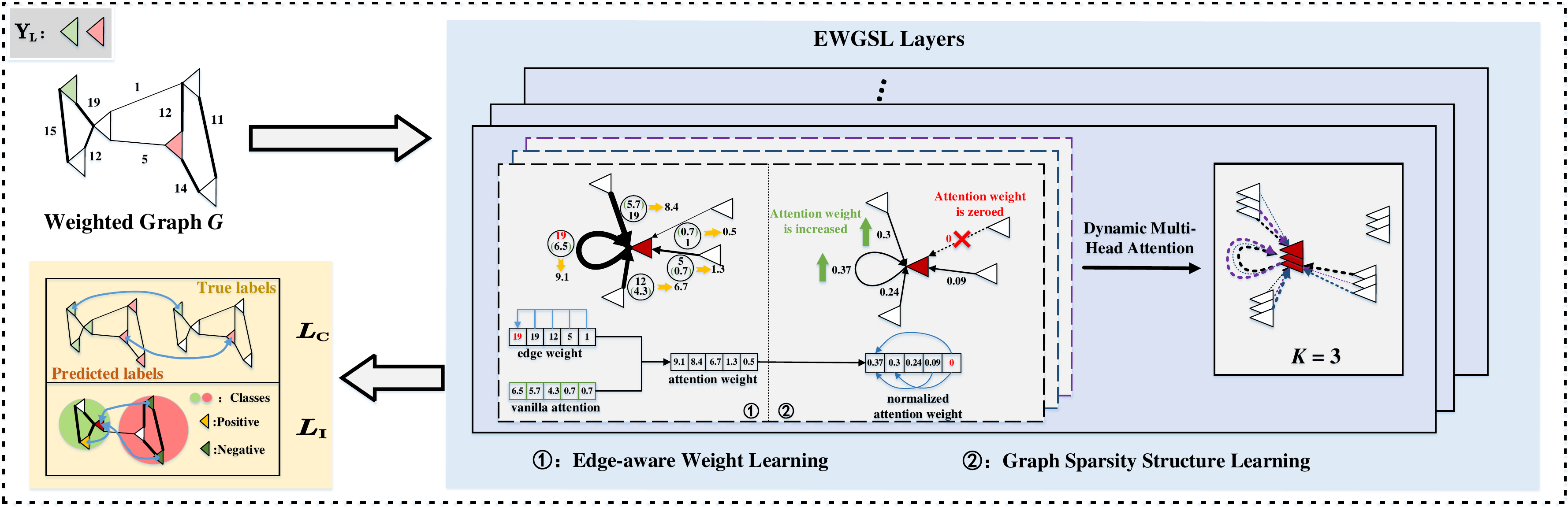}}
\caption{The architecture of EWGSL. The input consists of a weighted graph $G$ and a set of partially labeled nodes $Y_L$. Within the multiple layers of the EWGSL structure, as shown in the green box, node representations are progressively learned. Specifically, EWGSL incorporates edge weight information when calculating attention scores and employs the $\alpha$-entmax function for normalization. This process accurately captures the relationships between nodes by removing edges with low attention and reallocating attention to relevant nodes. This process is repeated multiple times (three times), ultimately using multi-head attention to learn node representations. The final loss function comprises two parts: $L_C$ and $L_I$.} \label{model}
\captionsetup{font=small}
\vspace{-0.5cm}
\end{figure*}

\vspace{-0.1cm}
\subsection{Problem Definition}
\vspace{-0.1cm}
\noindent\textbf{Weighted Graph:} A weighted graph is defined as $\mathrm{G=(V,E,\mathcal{W})}$, $\mathrm{V}=\{v_1,v_2,\cdots,v_n\}$ is the set of nodes, where $n$ is the number of nodes. $\mathrm{E}$ is the set of edges. $\mathcal{W}$ is the weight matrix. If there is an edge between $v_i$, $v_j$, $w_{ij}\neq0$, otherwise $w_{ij}=0$.

\noindent\textbf{Problem Formulation:} Given a weighted graph $\mathrm{G}$, and a small set of nodes $\mathrm{V_L}$ with labels $\mathrm{Y_L}$, the goal of the node classification task is to predict unlabeled nodes $\mathrm{V_U}$ with $\mathrm{Y_U}$, where ${\mathrm{V_L}}\cup{\mathrm{V_U}} = \mathrm{V}$.

\vspace{-0.1cm}
\subsection{Edge-aware Weight Learning}
\vspace{-0.1cm}
Edge-aware weight learning redefines the calculation of attention coefficients in GAT. Unlike vanilla GAT, which calculates attention coefficients based on node feature similarity, our approach incorporates edge weights into the computation. 

Larger edge weights signify stronger node connections in a weighted graph $\mathrm{G}$. Therefore, we can redefine the attention coefficients as follows:
\begin{equation}
\vspace{-0.2cm}
e_{ij}=\rho_{ij} \cdot {\boldsymbol{a}(\textbf{W}{\vec{h}_i},\textbf{W}{\vec{h}_j})},
\label{CW}
\vspace{-0.2cm}
\end{equation}
where $\textbf{W}$ and $\boldsymbol{a}$ represent the shared weight matrix and the shared mechanism used for computing the product between nodes. $\vec{h}_i$ and $\vec{h}_j$ represent the feature vectors of the nodes. $\rho_{ij}$ is the impact factor derived from the edge weights. 
 \begin{equation}
\rho_{ij}=\frac{w_{ij}}{\sum\limits_{k\in{{\mathcal{N}_i}}}w_{ik}}\;(i\notin{\mathcal{N}_i}),
\vspace{-0.2cm}
\end{equation}
where $w_{ij}$ denotes the weight between node $i$ and node $j$, ${\mathcal{N}_i}$ denotes the set of neighboring nodes for node $i$. 
\begin{equation}
w_{ii}=\mathrm{max}\ {\{w_{ij}\}}\;(j\in{\mathcal{N}_i},\ i\neq{j}).
\vspace{-0.2cm}
\end{equation}

When setting the weight of self-loops for nodes, we attempt three scenarios: taking the maximum, minimum, and average of the weights between the node and its neighboring nodes. Experimental results indicate that the best performance is achieved when taking the maximum value. 

This method captures feature similarity and the influence of edge weights, providing a more accurate depiction of relationships between nodes. Even when edge weights are small, if node features are highly similar, this approach can correct anomalous edge weights, ensuring that the calculation of attention coefficients is not significantly affected.

\vspace{-0.1cm}
\subsection{Graph Sparsity Structure Learning}
\vspace{-0.1cm}
Graph sparsity structure learning employs a novel activation function to normalize attention coefficients. Vanilla GAT uses the softmax function to assign non-zero weights to each neighboring node. However, this leads to difficulties in distinguishing true neighbors when there are noisy edges.

Compared to softmax, $\alpha$-entmax~\cite{ref16} offers greater flexibility by improving the softmax function through a hyperparameter $\alpha$. When $\alpha=1$, $\alpha$-entmax reduces to the original softmax function. The softmax function implements a standard probability distribution, resulting in relatively uniform weight distribution. For $1<\alpha<2$, the larger weights are further amplified, while smaller weights are weakened or even set to zero, resulting in sparsification. This sparsification enables the model to focus on key nodes or features while suppressing irrelevant information. Specifically, this can be expressed as:
\begin{equation}
{e}_{ij}^{\prime}=\mathrm{{\alpha}\text{-}entmax}({e_{ij}})={[(\alpha-1){e_{ij}}-\tau]_+^{1/{\alpha-1}}},
\label{SW}
\vspace{-0.2cm}
\end{equation}
where $[(\alpha-1){e_{ij}}-\tau]_+=\mathrm{max}\{0,\; [(\alpha-1){e_{ij}}-\tau]\}$, while $\tau$ denotes the threshold value for $\alpha$-entmax. When the attention coefficient satisfies ${e_{ij}}\leq\tau/{\alpha-1}$, the normalized probability becomes zero.

It is well known that $\alpha$-entmax must satisfy the condition that the sum of probabilities equals 1. Therefore, $\tau$ must satisfy:
\begin{equation}
\sum\limits_{j\in{{\mathcal{N}_i}}} \left[ (\alpha-1) e_{ij} - \tau(\vec{e}_i) \right]^{1/(\alpha-1)}_+ = 1,
\vspace{-0.2cm}
\end{equation}
where $\vec e_i$ represents the vector of attention weights for all edges connected to node $i$. 

There are two common methods for determining threshold functions: one is based on sorting~\cite{ref17}, and the other is bisection-based~\cite{ref18}. Taking into account the scenario where each node has a large number of neighboring nodes, the second method is selected to reduce computational efforts.

After obtaining the normalized attention coefficients, the feature vectors of the nodes can be updated by a linear combination of the neighboring node features and a nonlinear activation function:
\vspace{-0.2cm}
\begin{equation}
{\vec{h}_i}^{\prime}=\sigma({\sum_{j\in{{\mathcal{N}_i}}}}{{e}_{ij}^{\prime}\textbf{W}\vec{h}_j}).
\vspace{-0.25cm}
\label{R}
\end{equation}

The above describes the computation process of a single attention head. Multi-head attention mechanisms typically combine information from multiple heads through concatenation or averaging. However, these methods uniformly aggregate the data, which may overlook critical information in certain heads.

To dynamically adjust the weights of different heads based on input and task, we use learnable weight parameters $\Delta$ to allocate a weight for each head. Simultaneously, define a function $\mathrm{F}(\cdot)$ for aggregating multi-head information, and it is ultimately represented
as:
\begin{equation}
{\vec{h}_i}^{\prime} = \mathrm{F}(\Delta {\vec{h}_{i}}^{\prime}) = \mathrm{F}(\beta_k {\vec{h}_{ki}}^{\prime}) \quad (k = 1, 2, \ldots, K), 
\label{TR}
\vspace{-0.2cm}
\end{equation}
where $\beta_k$ represents the weight allocated to the $k$-th head in $\Delta$. Here, we employ a straightforward initialization method, obtaining $\Delta$ through multiple training iterations, and $\mathrm{F}(\cdot)$ is used for the average aggregation of information from multiple sources. Furthermore, various approaches to define $\Delta$ and $\mathrm{F}(\cdot)$ exist, and we will continue to explore them in our future work.

After obtaining the vector representations for each node, they are concatenated together to form the node representation matrix $\mathcal{H}$ for the entire graph:
\begin{equation}
\mathcal{H}=[{\vec{h}_1}^{\prime},{\vec{h}_2}^{\prime},\cdots,{\vec{h}_n}^{\prime} ],
\vspace{-0.2cm}
\end{equation}
where the size of $\mathcal{H}$ is $n\times{c}$, where $c$ represents the output dimension of each attention head and the number of classes.

\vspace{-0.1cm}
\subsection{Loss Function}
\vspace{-0.1cm}
Finally, we discuss the computation method of node labels and the loss function of the entire model.

We normalize the node representation matrix $\mathcal{H}$ to obtain the node membership matrix $\mathcal{M}$, where $m_{ij}$ represents the probability of node $i$ being assigned to class $j$.
\begin{equation}
\mathcal{M}=[\mathrm{softmax}({\vec{h}_1}^{\prime}),\mathrm{softmax}({\vec{h}_2}^{\prime}),\cdots,\mathrm{softmax}({\vec{h}_n})].
\vspace{-0.2cm}
\end{equation}

The class labels $\hat y_i$ of each node $i$ are obtained by maximizing its membership degree vector. Therefore, the inference function is:
\begin{equation}
\hat y_i={\mathrm{{\underset{{\textit{k}}\in\{1,2,\cdots,c\}}{argmax}}}m_{ik}}.
\label{cl}
\vspace{-0.2cm}
\end{equation}

Based on the known labels of some nodes and the learned node representations, the model measures the difference between the predicted and true labels of these nodes using a cross-entropy loss function.
\vspace{-0.1cm}
\begin{equation}
L_C= - \sum_{i \in V_L} \sum_{j=1}^{c} y_{ij} \log ( m_{ij}),
\vspace{-0.2cm}
\end{equation}
where $y_{ij}$ represents the true label of node $i$ for class $j$. If the true label of node $i$ is $j$, then $y_{ij}=1$; otherwise, $y_{ij}=0$.

\begin{algorithm}[t]
        % \vspace{-0.07cm}
        \fontsize{7}{12}\selectfont
	\caption{The pseudo-code of EWGSL.}
        \captionsetup{font=footnotesize}
	\label{algorithm1}
	\KwIn{ the adjacency matrix $\bf{\mathrm{\mathcal{W}}}$, node labels $Y_L$, and the hyperparameters $\alpha$, $\eta$, and $K$}
	\KwOut{ the predicted labels  $Y$, the total loss $L$, and the trained model }  
	\BlankLine
         Initialize the node representations matrix $\mathcal{H}$;	
	
 \While{convergence criterion is not satisfied}{
          
Update node representations $\mathcal{H}$ via Eq. (\ref{TR});

\While{K}{
    Learn attention weights via Eq. (\ref{CW});
    
    Sparsify attention weights via Eq. (\ref{SW});
    
    Update node representations via Eq. ({\ref{R}});
    }
    
    Update node labels via Eq. (\ref{cl});      	
    
    Compute the total loss $L$ via Eq. (\ref{L});
    
    Backpropagate and update model parameters using $L$;
	}
\end{algorithm} 

To further enhance the quality of node representations, enabling the model to not only train on label information but also fully leverage the implicit structural information within the graph, we introduce the InfoNCE loss. 

The goal of InfoNCE is to maximize the similarity of positive sample pairs while minimizing the similarity of negative sample pairs. Therefore, we can sample nodes that share the same class as the target node as positive samples, while sampling nodes from different classes across the entire node set as negative samples.
\begin{equation}
\vspace{-0.25cm}
L_{I} = -\frac{1}{n}\sum_{i=1}^{n}\log \frac{\lambda_{p} \exp\left({\text{sim}({\vec{h}_i}^{\prime}, {\vec{h}_j}^{\prime})}/{t}\right)}{{\sum\limits_{k \in N_i}} \lambda_{n} \exp\left({\text{sim}({\vec{h}_i}^{\prime}, {\vec{h}_k}^{\prime})}/{t}\right)},
\end{equation}
where $t$ denotes the temperature parameter, and $N_i$ represents the set of negative samples of node $i$. To account for the impact of weights, we use the average attention weight within the class (i.e., $\lambda_{p}$) as the weight factor for positive samples and the average attention weight between classes (i.e., $\lambda_{n}$) as the weight factor for negative samples. 
\vspace{-0.15cm}
\begin{equation}
\lambda_{p}=\frac{1}{|\mathcal{E}_{\text{intra}}|} \sum{ {e}_{ij}^{\prime} \; (i \neq j, \;\hat y_i = \hat y_j}),
\vspace{-0.15cm}
\end{equation}
\begin{equation}
\lambda_{n}=\frac{1}{|\mathcal{E}_{\text{inter}}|} \sum{{e}_{ij}^{\prime} \; (i \neq j, \;\hat y_i \neq \hat y_j}),
\vspace{-0.15cm}
\end{equation}
where $\mathcal{E}_{\text{intra}}$ and $\mathcal{E}_{\text{inter}}$ denote the sets of edges between nodes with the same labels and different labels, respectively.

Our model aims to minimize the loss function presented below as its primary objective:
\vspace{-0.1cm}
\begin{equation}
L = L_C+\eta{L_I},
\label{L}
\vspace{-0.1cm}
\end{equation}
where $\eta$ is the balance parameter that controls the loss of $L_I$.

\begin{table*}[t]
\vspace{-0.3cm}
\centering
\fontsize{6.5}{11}\selectfont
\caption{The results of node classification on datasets.}
\vspace{-0.2cm}
\captionsetup{font=footnotesize}
\begin{tabular}{c?c|c|c|c|c?c|c|c|c|c|c|c}
\specialrule{0.2pt}{\abovetopsep}{0pt}
\bottomrule
\diagbox[width=1.9cm]{Dataset}{Method} & {Metric} & GAT & GATv2 & GCNII & SAGE & SUBLIME& MetaGC  & SSGCN & PTDNet & PROSE & SGAT & EWGSL \\
\specialrule{0.2pt}{\abovetopsep}{0pt}
\hline
 \multirow{2}{*}{\raggedleft Vessel2015-01} &\!\!ACC & 59.3  & 67.8  & 55.7 & 72.9 & 31.7 &37.9 &54.3 & 58.5  & 65.7 & \underline{77.2} & \bf{81.5} \\ 
 & F1  & 65.4  & 69.1 & 57.8 & 76.4 & 20.4 &38.1 &27.3 & 70.4 & 22.0 & \underline{83.2} & \bf{85.3} \\ 
\hline
\multirow{2}{*} {\raggedleft Vessel2015-10}&\!\!ACC & 60.7  &68.1 &57.2  & 73.1 & 32.0  &30.0 &47.2 & 56.4 & 67.1 & \underline{77.0} & \bf{77.9}  \\
 & F1  &64.2 & 74.6 & 56.4 & 71.8& 21.1 &32.5 &24.4 & 68.7 & 21.4 & \underline{80.7} & \bf{82.6} \\ 
\hline
\multirow{2}{*} {\raggedleft Vessel2018-10}&\!\!ACC & 58.6 &67.5  & 53.1 & 74.8 & 31.7 &31.4 & 31.7 & 56.8 & 67.8 & \underline{76.8} & \bf{79.3} \\
 & F1  & 67.3  & 72.4 & 55.5 & 74.7& 37.4 &36.3 &43.3 & 66.4 & 29.0 & \bf{86.9} & \underline{84.9} \\ 
\hline
\multirow{2}{*} {\parbox[c][0.7cm][c]{1.4cm}{\centering ML-100k$\_$ES \\ \vspace{-0.5ex} N = 0$\%$}} &\!\!ACC &33.7 &39.6  &28.6 &\underline{42.0} &20.8 &20.9 &34.9 &40.5 &41.4 &41.4 &\bf{45.0} \\
 & F1  & 48.6 &\underline{51.3}& 30.4 & 45.6 & 12.3 &21.1 &14.4 & 40.0 & 26.9 & 47.5 & \bf{60.7} \\ 
\hline
\multirow{2}{*} {\parbox[c][0.7cm][c]{1.4cm}{\centering ML-100k$\_$ES \\ \vspace{-0.5ex} N = 5$\%$}}&\!\!ACC &32.4 &39.2 &27.4  &42.9 &20.1 &23.1 &33.3 &41.2 &\underline{44.2} &41.9 &\bf{44.3}\\
 & F1  & 48.2 &51.1 &30.9 & \underline{48.7} & 11.2 &22.3 &14.7 & 40.8 & 27.1 & 45.1 & \bf{61.2} \\ 
\hline
\multirow{2}{*} {\parbox[c][0.7cm][c]{1.4cm}{\centering ML-100k$\_$ES \\ \vspace{-0.5ex} N = 10$\%$}}&\!\!ACC  &31.4 &38.3 &27.0  &41.3 &22.0 &25.4 &32.5 &41.2 &41.6 &\underline{42.0} &\bf{44.9}  \\
 & F1  & 44.1 &\underline{49.8} &28.5  & 46.1 & 11.6 &19.1 &14.9 & 47.0 & 26.9 & 49.5 & \bf{58.8} \\ 
\hline
\multirow{2}{*} {\parbox[c][0.7cm][c]{1.4cm}{\centering ML-100k$\_$ES \\ \vspace{-0.5ex} N = 15$\%$}} &\!\!ACC &31.0 &36.7  & 27.1&40.2 &22.0 &26.3 &32.1 &\underline{42.0} & 39.5&41.5 &\bf{42.4} \\
 & F1  & 43.6 &48.1 &28.6 & 45.6 & 12.6 &17.8 &14.9 & 41.1 & 24.6 & \underline{48.7} & \bf{58.2} \\ 
\specialrule{0.2pt}{\abovetopsep}{0pt}
\bottomrule
\end{tabular}
\vspace{-0.3cm}
\label{table2}
\end{table*}

\begin{table}[t]
\begin{center}
\vspace{-0.2cm}
\fontsize{7}{12}\selectfont
\caption{Information regarding datasets.}\label{table1}
\captionsetup{font=footnotesize}
\begin{tabular}{c|r|r|c|c}
\specialrule{0.2pt}{\abovetopsep}{0pt}
\bottomrule
\textbf{Dataset} & \multicolumn{1}{c|}{\textbf{$\#$Nodes}} & \multicolumn{1}{c|}{\textbf{$\#$Edges}} & \textbf{$\#$Classes} & \textbf{Density} \\ 
\specialrule{0.2pt}{\abovetopsep}{0pt}
\hline
Vessel2015-01        & 767                                 & 130,153                            & 5                    & 0.442            \\
Vessel2015-10        & 986                               & 299,055                             & 7                    & 0.615            \\
Vessel2018-10        & 401                                 & 61,308                              & 3                    & 0.761            \\
ML-100k$\_$ES          & 1,612                               & 58,439                              & 9                    & 0.045            \\ 
\specialrule{0.2pt}{\abovetopsep}{0pt}
\bottomrule
\end{tabular}
\end{center}
\vspace{-0.6cm}
\end{table}

The pseudo-code of our method is listed in Algorithm~\ref{algorithm1}. First, the node representations are initialized and refined through the EWGSL model to obtain more precise node embeddings. Within the multi-head attention mechanism, each attention head is iteratively processed to compute attention coefficients and update the node embeddings. Finally, the node labels and loss function are updated iteratively until the maximum number of iterations is reached or the loss function stabilizes over several consecutive iterations, satisfying the convergence criteria. Once these conditions are met, the algorithm terminates.

\section{Experiments}
\vspace{-0.1cm}
\subsection{Experiment Settings}
\vspace{-0.1cm}
\noindent\textbf{Datasets:} We conduct experiments on four datasets, and the information for each dataset is presented in TABLE ~\ref{table1}. 

The vessel datasets (i.e., Vessel2015-01, Vessel2015-10, and Vessel2018-10) consist of vessel trajectory data collected over a month in four cities. The collection of these vessel trajectory record information is accomplished with the help of the Vessel Monitoring System. During the operation of the system, various sensors are in a continuous working state, capturing the status information of the vessels in real-time and transmitting it to the data processing center in a timely manner. These continuous streams of raw data are finally constructed into a complete vessel dataset after a series of processing and integration processes.
\begin{itemize}

\item When the spherical distance between two vessels is less than 1.11 kilometers, they are considered to meet, creating an edge with a weight increment of one.

\item As vessels in the same county frequently encounter each other, we use the county to which a vessel belongs as the true label.

\end{itemize}

ML-100k$\_$ES\footnote[1]{https://github.com/wtt2022/DataSet/} is constructed from the open dataset MovieLens 100K\footnote[2]{https://grouplens.org/datasets/movielens/}. The dataset includes multiple files, and we focus on u.data and u.item. We will release this dataset.

\begin{itemize}
\item Construct a graph from the rating records in u.data by filtering and sorting the ratings by timestamp. Compute edge weights based on consecutive movie ratings: if two movies are rated consecutively, increase the edge weight between those movies by 1.

\item Assign labels to movies based on genres in u.item. As movies can belong to multiple genres, we determine the label for each movie by assigning the genre with the highest frequency across the dataset.

\end{itemize}

\noindent\textbf{Baselines:} 
We select six state-of-the-art (SOTA) methods for node classification and categorize them into three groups based on how they handle noise: 1) Weight learning methods: SUBLIME\cite{ref14}, MetaGC\cite{refer7}; 2) Structure learning methods: SSGCN\cite{ref15}, PTDNet\cite{refer5}, PROSE\cite{refer11}; and 3) Both: SGAT~\cite{refer6}. In addition, we consider four SOTA Graph Neural Network models or other sparse methods: GAT~\cite{refer4}, GATv2~\cite{refer22}, GCNII~\cite{refer23}, SAGE~\cite{refer24}.

\noindent\textbf{Implementation:} The first three datasets contain noise due to human interventions during their construction, such as trajectory data collection, timestamp alignment, and vessel encounter identification. In our experiments, we introduce random noise edges into the fourth dataset at varying proportions, denoted by the variable ``N''. Specifically, we randomly add 5$\%$, 10$\%$, and 15$\%$ of edges to ML-100k$\_$ES to create new experimental datasets. 

We use a three-layer GAT to learn node representations, with hidden layers of sizes 256 and 128, and input and output sizes of $n$ and $c$, respectively. The model trains for 100 epochs with a learning rate of 0.005. Based on the dataset construction, we divide the datasets into two categories for hyperparameter experiments. We search the hyperparameters $\alpha$, $K$, and $\eta$ within the ranges \{1,\ldots,2\}, \{2,\ldots,12\}, and \{0,\ldots,0.5\}, respectively. The results show that the model performs best when $\alpha$, $K$, and $\eta$ are set to \{1.5, 1.5\}, \{6, 8\}, and \{0.05, 0.1\}, respectively. We employ the Adam optimizer to minimize the total loss.

To enhance generalization and reduce randomness effects, we conduct multiple experiments with different random seeds and average the results. 

\noindent\textbf{Metrics:} We evaluate the accuracy of a node classification by two common metrics Accuracy (ACC) and Micro-F1 (F1), and a higher score means a more accurate result.
\vspace{-0.1cm}
\subsection{Experment Results}
\vspace{-0.1cm}
The experimental results are shown in TABLE \ref{table2}, with the best results highlighted in bold and the second-best results underlined. As seen from the table, EWGSL shows an average improvement of 17.8$\%$ in F1 and 3.4$\%$ in ACC compared to the best baseline.

The results demonstrate that graph structure learning is superior to weight learning methods in handling noisy graphs, indicating that graph structure learning has a greater advantage in identifying and filtering out noisy edges. Compared to merely adjusting weights, optimizing and refining the graph structure more effectively enhances model performance, enabling a more accurate representation of the true structure and relationships within the graph.

Moreover, although the performance of SGAT is close to that of EWGSL, it does not surpass it. This is primarily because SGAT assigns a unique attention coefficient to each edge, limiting its global information capture and adaptability to context changes. As a result, SGAT loses some details when processing intricate information.

\vspace{-0.1cm}
\subsection{Ablation Study}
\begin{figure}[b]
\vspace{-0.7cm}
\centering
\begin{minipage}{0.235\textwidth}
  \centering
  \vspace{0.2cm}
  \includegraphics[width=1.07\textwidth, height=0.9\textwidth, keepaspectratio=false]{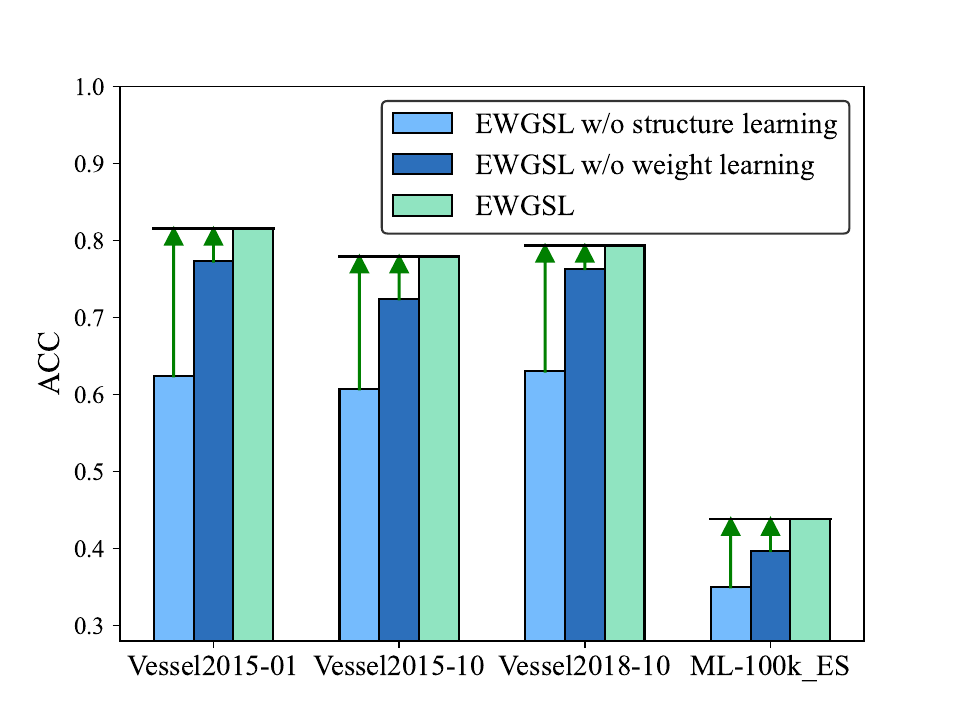} % Use width instead of scale
  \caption{Abiation study results.}
  \captionsetup{font=footnotesize}
  \label{ab}
\end{minipage}
\hspace{0.01\textwidth} % Adjust the spacing between the figures
\begin{minipage}{0.23\textwidth}
  \centering
  \includegraphics[width=1\textwidth, height=1\textwidth, keepaspectratio=false]{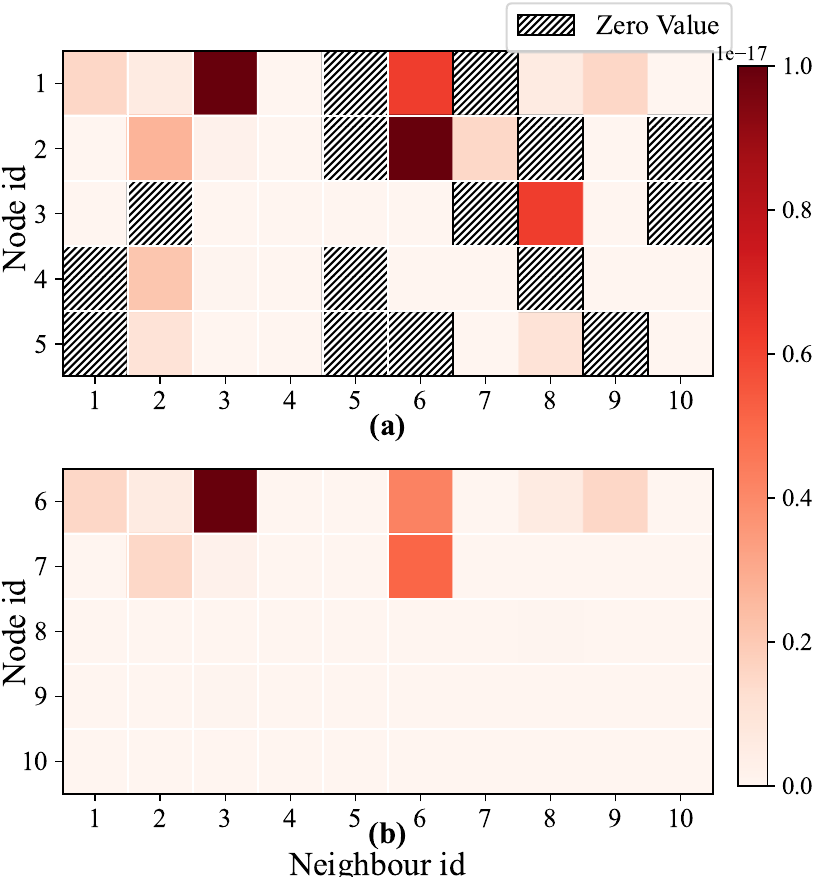} % Use width instead of scale
    \vspace{-0.6cm}
  \caption{Case study results.}
  \captionsetup{font=footnotesize}
  \label{ca}
\end{minipage}
\vspace{-0.6cm}
\end{figure}

\vspace{-0.1cm}
We conduct ablation experiments on four datasets to evaluate the effectiveness of the two key modules in EWGSL. For instance, as shown in Fig.\ref{ab}, the ACC of EWGSL increases by an average of 26.57$\%$ compared to models using only the edge-aware weight learning module, and by 6.82$\%$ compared to models using only the graph sparsity structure learning module. This indicates that graph sparsity structure learning significantly enhances model performance. Specifically, this module more effectively optimizes graph structure and accurately captures node relationships with noise.

Moreover, the impact of the graph sparsity structure learning module on the ML-100k$\_$ES dataset is smaller compared to other datasets, indicating that its advantages are more pronounced when handling denser graph data. Specifically, the graph structure of the ML-100k$\_$ES dataset is relatively sparse, sothe advantage of optimizing graph structure and capturing node relationships is less significant than in denser data. In contrast, the node relationships in dense graph data are more complex, and the graph sparsity structure learning approach can improve the graph structure more effectively, thereby enhancing the performance of the model.
\vspace{-0.1cm}
\subsection{Case Study}
\vspace{-0.1cm}
To comprehensively evaluate the effectiveness of EWGSL in handling noisy graphs, we randomly selected five nodes from the ML-100k\_ES dataset and chose ten neighboring nodes for each. We then computed the attention weights between these nodes and their neighbors using both GAT and EWGSL.

The heatmap in Fig.~\ref{ca} illustrates a comparison of the attention weights calculated by GAT and EWGSL. By comparing the results from both methods, it is clear that EWGSL can more effectively adjust the distribution of attention, enhancing the expressiveness of the graph structure. As shown in Fig.~\ref{ca}~(b), Node 2 has strong connections with two neighboring nodes, and after applying EWGSL, the weights of these two neighbors significantly increase, with their colors deepening, indicating that their importance within the graph is more strongly highlighted (see Fig.~\ref{ca} (a)). Moreover, EWGSL is also capable of effectively identifying and suppressing the influence of noise. In this experiment, EWGSL successfully identified three irrelevant nodes with weak relationships to Node 2 and its neighbors, effectively reducing their weights and mitigating their interference with the graph representation. In contrast, the traditional GAT performed poorly in handling these irrelevant nodes, failing to effectively distinguish between important nodes and noise. This result demonstrates that EWGSL can better optimize the graph structure when dealing with noisy graph data, thereby improving the performance of the model.

\section{Conclusion}

In this paper, we introduce a novel model aimed at addressing node classification challenges in noisy weighted graphs. Our model innovatively integrates edge weights into the attention mechanism of the GAT, providing a finer-grained representation of the graph structure. Additionally, we introduce a sparse graph structure that employs the $\alpha$-entmax function to filter out noisy edges and emphasize relevant connections, thereby enhancing the accuracy and reliability of the learned graph structure. Extensive experiments show that EWGSL significantly surpasses SOTA methods.

After completing the current research, our future research will focus on addressing the challenges posed by the rapid growth of data volume. As the data volume surges sharply, the computational complexity of large weighted graphs increases dramatically, and the traditional computing resources and algorithmic efficiency have revealed significant limitations. Therefore, it is of utmost urgency to optimize existing algorithms and explore new computing architectures to reduce the computational complexity.

\section*{Acknowledgment}

This research is supported in part by the National Science Foundation of China (No. 62302469), the Natural Science Foundation of Shandong Province (ZR2023QF100, ZR2022QF050).

\bibliographystyle{IEEEtran}
\bibliography{confer}

\end{document}